\documentclass[runningheads]{llncs}
\usepackage{graphicx}
\usepackage{amsfonts}
\usepackage{amsmath}

\usepackage[
backend=biber,
style=numeric,
sorting=ynt
]{biblatex}
\begin{document}
\title{Behaviour-conditioned policies for cooperative reinforcement learning tasks}
%
%\titlerunning{Abbreviated paper title}
% If the paper title is too long for the running head, you can set
% an abbreviated paper title here
%
\author{Antti Keurulainen\inst{1,3} \and
Isak Westerlund\inst{3} \and
Ariel Kwiatkowski\inst{3} \and\\ Samuel Kaski\inst{1,2} \and Alexander Ilin \inst{1}}

\authorrunning{A. Keurulainen et al.}
% First names are abbreviated in the running head.
% If there are more than two authors, 'et al.' is used.
%

\institute{Helsinki Institute for Information Technology HIIT, Department of Computer Science, Aalto University
\and
Department of Computer Science, University of Manchester
\and
Bitville Oy, Espoo, Finland}

\maketitle              % typeset the header of the contribution
\begin{abstract}
The cooperation among AI systems, and between AI systems and humans is becoming increasingly important. In various real-world tasks, an agent needs to cooperate with unknown partner agent types. This requires the agent to assess the behaviour of the partner agent during a cooperative task and to adjust its own policy to support the cooperation. Deep reinforcement learning models can be trained to deliver the required functionality but are known to suffer from sample inefficiency and slow learning. However, adapting to a partner agent behaviour during the ongoing task requires ability to assess the partner agent type quickly. We suggest a method, where we synthetically produce populations of agents with different behavioural patterns together with ground truth data of their behaviour, and use this data for training a meta-learner. We additionally suggest an agent architecture, which can efficiently use the generated data and gain the meta-learning capability. When an agent is equipped with such a meta-learner, it is capable of quickly adapting to cooperation with unknown partner agent types in new situations. This method can be used to automatically form a task distribution for meta-training from emerging behaviours that arise, for example, through self-play.

\end{abstract}

\section{Introduction}

In many real-world applications, the ability of AI agents to cooperate with each other and humans is of crucial importance. It is especially important to have the capability to cooperate with different types of agents with different characteristics and behavioural patterns. In practice, this calls for functionalities to infer the behaviour of the partner agent and then adapting to the inferred agent type for common good. Furthermore, it is useful to infer the behaviour of an unknown partner agent as fast as possible during the ongoing task. Such situations arise, for example, when self-driving cars try to adapt to the various driving styles of the surrounding (human) drivers, or when an AI teacher tries to adapt to the skill level of the human student.

Human cognition studies suggest that predictions on other peoples' behaviour are based on high-level models, such as mental states and abstractions  \cite{premack1978does,gopnik1992child,baker2017rational}. Such high-level abstractions may be, for example, intentions, goals, desires or some other factors describing the mental state. Taking inspiration from human cognition, our goal is to enable an agent to infer the latent variable that explains the behaviour of the partner agent, and then make use of the inferred information during the ongoing task.  In practice, the latent variable needs to be low-dimensional since it needs to be inferred from short-term behaviour. For example, it might be known that the behaviour of the partner agent depends heavily on the skill level of the partner agent, which may be unknown and therefore needs to be inferred.

In our method, we synthetically produce partner agent populations that are used for training a meta-learner, which can be integrated as part of the agent policy. As a result, an agent gains the capability of "learning to learn" and can quickly adapt to unknown agent types in new situations. Self-play has been shown to create an automatic curriculum for learning and to produce emerging behaviours \cite{baker2019emergent,leibo2019autocurricula}, thus it is a good candidate to synthetically produce populations with different behavioural patterns. Since the agents are synthetically generated, the process also produces appropriate ground truth data of the partner agent behaviour that can be used for training the meta-learner.

We frame this setting as a deep meta-reinforcement learning problem \cite{duan2016rl,wang2016learning,stadie2018some}, where the partner agent behaviour is embedded in the transition dynamics of the Markov Decision Process (MDP). During the training phase of the meta-learning process, an agent is partnered with different partner agent types in different instances of the environment. A detailed description of the meta-learning approach is described in section 3.

Our method is also based on the assumption that the two tasks, of inferring the partner agent behaviour, and of conducting the actual task, have different requirements and hence need different solutions. In particular, the former requires observing the agent for some time period and hence requires memory in its policy. A common way for introducing memory in neural networks is by allowing recurrent connections, which brings challenges such as vanishing and exploding gradients and credit assignment \cite{bengio1994learning}. On the other hand, many cooperative tasks can be executed without memory, once the policy network is conditioned on the partner agent behaviour. As an example, if the partner agent is closer to a specific item to be collected in a gridworld, it is useful to let the partner agent collect the item, if it is known to have the skills to do it.

In our approach, which we call the \textit{behaviour-conditioned policy}, we suggest a separate dedicated network with memory that infers the agent behaviour quickly during the first steps of the task execution and a separate policy network without memory, which is conditioned on the inferred behaviour. As a result, the actual policy is easier to train as it can be implemented by a simpler feedforward network without recurrent connections. Both networks can be trained separately by using the synthetically produced agent populations together with the ground truth data of their behaviour. During the execution, the ground truth data is replaced by the predictions of the partner agent behaviour, resulting in a policy that is not dependent on the ground truth information that is used for training.

The contributions of this paper can be summarized as:

\begin{itemize}
  \item A method to automatically generate a task distribution and associated ground truth data from scratch by self-play for training a meta-learner.
  \item An architecture and training method that can efficiently use the generated task distribution and the ground truth data.
\end{itemize}

We demonstrate the capabilities of our approach in two different types of experiments. In the first experiment, we compare the performance against the $RL^2$ meta-reinforcement learning architecture \cite{duan2016rl} in a simple matrix game. In the second, more complex environment, an agent solves a travelling salesman gridworld (TSG) task in cooperation with a partner agent of an unknown type, and we show higher performance when compared to an end-to-end solution with an LSTM architecture.

\section{Related work}

There is a long history of research in opponent modelling \cite{albrecht2018autonomous}. For example, \cite{raileanu2018modeling} present an idea, where one agent predicts the behaviour of the other agent by putting itself in a similar situation and \cite{hu2020other} present a method of breaking the symmetries in the underlying task and thus improving cooperation. The former is an example where the partner agent is considered to be similar to oneself, and the latter assumes that the partner agent is optimal but might have converged to a different convention than oneself. In our case, we make weaker assumptions as the partner agent is not assumed to be optimal. In \cite{foerster2017learning} and \cite{letcher2018stable}, methods to shape the learning of the other agent are presented. In our work, we do not try to affect the learning of the partner agent, but rather adapt to the behaviour of the agent, which is embedded in the transition dynamics.

Applying centralized training with decentralized execution (CTDE) \cite{oliehoek2012decentralized,kraemer2016multi} is an active research area and it is an open problem how to best benefit from the centralized training when private information is not available during execution. For example, MADPPG \cite{lowe2017multi} uses a centralized critic with decentralized actors and \cite{foerster2018counterfactual} suggests a method based on using a centralised critic and a specific version of the baseline in an actor-critic algorithm to tackle the multi-agent credit assignment challenge. Since centralized critics appear to be beneficial, we make use of this concept in our baselines and thus reduce variance in the policy gradient estimates. Our method suggests a natural extension of CTDE to the transition dynamics of the MDP.

Self-play has been shown to be a successful method for producing an automated curriculum for learning in multi-agent reinforcement learning settings \cite{silver2016mastering,baker2019emergent,leibo2019autocurricula}. When training is conducted with a copy of itself, the cooperability with other agent types with different behaviour might be limited. One possible solution to improve generalization and robustness in self-play is to interleave old versions of the policy in the training procedure. In \cite{silver2016mastering}, a method to sample old opponents, which were stored during the self-play training, was used to prevent overfitting to the latest policy and for stabilizing the self-play training procedure. Also, \cite{bansal2017emergent} show that sampling opponents from a pool of past versions of the policy improves the performance and robustness of training agents in adversarial settings. Our method also benefits from the sampling of old opponents, but instead of adversarial training we use self-play to produce a suitable task distribution for meta-learning.

One typical way to encourage neural networks to learn more useful representations for a particular task, is to add auxiliary tasks to the learning process \cite{suddarth1990rule}. For example, auxiliary tasks can be used to improve representations in classification tasks \cite{rasmus2015semi,zhang2016augmenting}, or in the context of deep reinforcement learning \cite{mirowski2016learning,jaderberg2016reinforcement,hu2019simplified}. Our mechanism differs from the typical way of using auxiliary heads, as instead of adding an extra auxiliary head to the common representation, we train a distinct prediction network with a supervised loss, and condition the policy network with this prediction.

\section{Setting}

We formalize our problem as a meta-reinforcement learning setting \cite{duan2016rl,wang2016learning,stadie2018some}. We synthetically construct a set of MDPs by training populations of agents with self-play to obtain a distribution of agents with different behaviours. We sample MDPs from a family $\mathcal{M}$  according to a meta learning distribution $M_i \sim p(M)$. From the point of view of one agent, every MDP has a different transition dynamics dictated by the behaviour of the partnering agent. To be more specific, we define the MDP by the tuple $ M = \langle \mathcal{S}, \mathcal{A}, \mathcal{P}, \mathcal{R}, \gamma, \rho_0, H \rangle $, where $\mathcal{S}$ is the state space and $\mathcal{A}$ is the action space, $\gamma$ is the discount factor, $\rho_0 : \mathcal{S} \rightarrow [0,1] $ is the initial state distribution and $H$ is the horizon. The transition function $\mathcal{P}$ is defined as a function of the action space and the state space as $\mathcal{P} : \mathcal{S}  \times  \mathcal{A} \times \mathcal{S} \rightarrow [0,1]$ and $\mathcal{R}: \mathcal{S} \times \mathcal{A} \rightarrow {\rm I\!R}$ is the reward function experienced by the agent.

Under this setting, the objective is to maximize the expected sum of discounted rewards over the episodes $J(\pi_\theta) = \mathbb{E}_{\tau\sim p_\pi (\tau)}[\sum_{t=0}^{H}\gamma^t \mathcal{R}(s_t,a_t)]$, where $\tau = (s_0,a_0, ...)$ is the trajectory of states $s_t$ and actions $a_t$ of the agent at time step $t$ for an episode of length $H$. The initial state is sampled from the distribution $\rho_0$ and the agent samples the action $a_t$ from its policy function $\pi(a_t|s_t)$. The next state is sampled from the transition dynamics function $s_{t+1} \sim \mathcal{P}(s_{t+1}|s_t,a_t)$. The meta-MDP objective is to find the policy that maximizes the expected returns over the task distribution: $\text{arg}\,\max\limits_{\theta} \mathbb{E}_{p(M)}[J(\pi_\theta)]$.

\section{Behaviour-conditioned policy}

\begin{figure}
\includegraphics[width=1.0\textwidth]{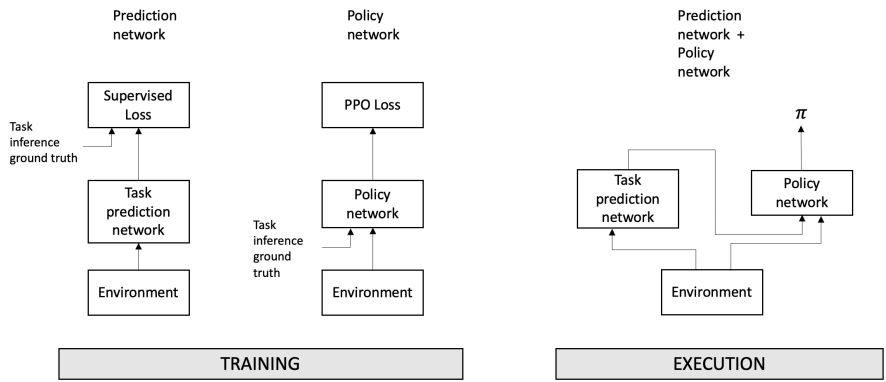}

\caption{Illustration of the training and execution architectures. Two separate networks are trained with ground truth data to predict the task and to run the policy. The execution is run by using the task prediction to allow decentralised execution}
\label{trainexec}
\end{figure}

The two distinct structures of the task can be efficiently exploited by constructing separate networks for both task structures, which we name the \textit{task prediction} and the \textit{policy}. We synthetically produce populations of agents together with ground truth information about their behaviour. Under the meta-learning framework, each sampled MDP embeds the behaviour of the partner agent, and the task prediction refers to predicting the behaviour of the partner agent. Following the centralised training and decentralised execution (CTDE) paradigm, we use this ground truth information during training, but not during decentralised execution. The architectures for the training and execution phases are shown in Fig \ref{trainexec}. 

The task prediction network is stateful and thus able to preserve the representation of the trajectories $\tau_i = \{s_t\}_{t=0}^l$, where $i$ is the task index and $l$ is the length of the episode. The task prediction is performed by a neural network $f_\phi$ as $\hat{T}_i = f_\phi (\tau_i) $, where $\hat{T_i}$ is the estimation of the task $i$. More specifically, the task prediction network uses a standard LSTM architecture \cite{hochreiter1997long}. The task prediction network is trained by minimizing the loss under the task distribution $\mathcal{L} = \mathbb{E}_{p(M)}[L_M(f_\phi)]$ by using the ground truth task labels ${T}_i$. 

During the training, the policy network is a feedforward network $\pi = g_\theta (s_i,T_i)$. During the execution, the ground truth task label ${T}_i$ is replaced by the prediction $\hat{T}_i$, produced by the task prediction network.

\section{Experiments}

In this section, we present the experiments and their results. In these experiments one agent (\textit{the main agent}) is equipped with the behaviour-conditioned policy, and adapts to the behaviour of the other agent (\textit{the partner agent}). The first experiment is a simple matrix game, where the task is to adapt to the partner policy. The purpose of this experiment is to investigate how well the suggested architecture can benefit from a synthetically generated task distribution compared to a strong meta-learning baseline. In the second experiment, we combine the behaviour conditioned policy architecture with the creation of a meta-learning task distribution with self-play. In essence, the task is a travelling salesman-type problem for which self-play produces a distribution of agents of different skill levels.

\subsection{Matrix game experimental setup}

In this matrix game two agents perform actions sampled from their action distributions. For each time step, the shared reward of their joint action is defined by the payoff matrix, shown in Table \ref{pm_matrix}.

For the partner agent, the action distributions are generated by drawing them from a symmetric Dirichlet distribution $Dir(\alpha)$ with concentration parameter $\alpha$. The main agent is equipped with the behaviour-conditioned policy and it is thus able to learn during the episode the behaviour of the unknown partner agent, and adjust its own policy accordingly.

One episode is fixed to ten steps, and the partner agent uses the same sampled action distribution throughout the episode, and for each episode a new action distribution is sampled. In one training iteration, 50 partner action distributions are drawn and thus one iteration consists of 50 different tasks in the meta-training scheme. During the training, the task prediction network of the main agent learns to predict the task (the action distribution of the partner agent) by using ground truth information, and the policy network learns to conduct the task by using ground truth information of the action distribution. During execution, the ground truth information is replaced by the predictions from the task prediction network.   

The difficulty of the task can be adjusted by altering the concentration parameter $\alpha$. Small values of $\alpha$ create a high probability of one single action and make the prediction of the partner agent behaviour easy. Higher values of $\alpha$ create action distributions that are more even, in which case predicting the partner agent action distribution is more difficult.

The payoff matrix is designed in such a way, that the better the interpretation of the other agent behaviour, the higher the payoff. As an example, predicting the partnering agent to take an action $p_0$ results in high reward if predicted correctly, and a high negative reward if predicted wrongly. If the prediction of the partner agent behaviour is uncertain, the main agent is tempted to select the action $m_4$, a low risk but low reward option, to avoid high negative rewards.

\begin{table}[t]
\begin{center}
\begin{tabular}{ c|c|c|c|c|c| } 
&$m_0$ & $m_1$ & $m_2$ & $m_3$ & $m_4$\\
 \hline
$p_0$&1.0 & -0.7 & -0.4 & -0.1 & 0.0\\
 \hline
 $p_1$&-1.0 & 0.8 & -0.4 & -0.1 & 0.0\\
 \hline
 $p_2$&-1.0 & -0.7 & 0.6 & -0.1 & 0.0\\
 \hline
 $p_3$&-1.0 & -0.7 & -0.4 & 0.4 & 0.0\\
 \hline
 $p_4$&-1.0 & -0.7 & -0.4 & -0.1 & 0.2\\
 \hline

\end{tabular}
\end{center}
\caption{The payoff matrix for the matrix game. The partnering agent actions represent the rows $p_0-p_4$, and the main agent selects the columns $m_0-m_4$. }
\label{pm_matrix}
\end{table}

The prediction network uses the MSE loss function $\mathcal{L} = \mathbb{E}_{p(M)} \|T-\hat{T}\|_2^2$ and the policy network is trained using PPO loss \cite{schulman2017proximal}, with separate networks for policy and value predictions. In this experiment the input is augmented by the time step index, in order to allow the network to perform value prediction. More details of the network implementation is available in the appendix.

\subsection{Matrix game results}

We compare our algorithm against the $RL^2$ meta-reinforcement learning algorithm \cite{duan2016rl}, which is capable of performing end-to-end training in the matrix game task. The expressive capacity of the behaviour conditioned policy was designed to be similar to the baseline $RL^2$ implementation, see details about the specific architectures in the appendix. Both methods receive the ground truth information of the partner agent behaviour during training by the use of a centralized critic, in order to reduce the variance of the policy gradient estimates.

\begin{figure}[t]
\centering
\includegraphics[width=0.9\textwidth]{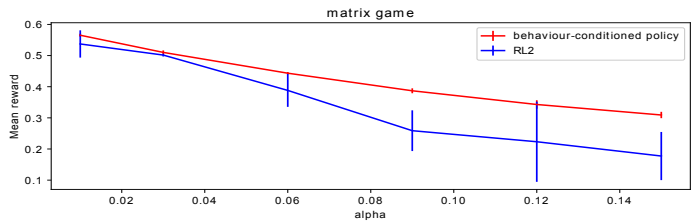}
\caption{Results of the matrix game experiment. The behaviour-conditioned policy is able to learn to predict the partner agent behaviour during an episode, and use that information in its policy function efficiently. The results show that the baseline $RL^2$ method does not reach as high mean reward, and the variance across the seeds is considerably higher. Lines are mean values over five random seeds and error bars indicate the standard deviation. The rewards are measured at the last time step of the episodes. }
\label{matrix_game}
\end{figure}

In order to compare the performance for various levels of task difficulty, the experiment is run with six different values of the concentration parameter $\alpha$. When $\alpha=0.01$ or $\alpha=0.03$, the task is easy and can be solved by both methods, but as the task is made more difficult by increasing the value of $\alpha$, the behaviour-conditioned policy outperforms the end-to-end alternative (Fig. \ref{matrix_game}). The results show that if the factor that explains the behaviour is known, a separate network can infer the partner agent behaviour fast enough with sufficient accuracy during an episode, and that conditioning a simple feedforward policy with the inferred behaviour reaches higher mean reward and more stable training compared to the baseline, as is shown in Fig \ref{matrix_game}.

\subsection{Travelling Salesman Gridworld (TSG) experimental setup}

\begin{figure}[t]
\centering
\includegraphics[width=0.3\textwidth]{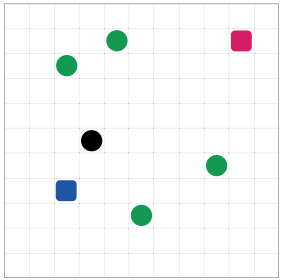}

\caption{Illustration of the 11x11 gridworld. Red and blue squares are the two agents, black circle is the final goal, green circles are the subgoals.}
\label{gridworld}
\end{figure}

The TSG experiment is designed to encourage cooperation and quick task inference during the execution. In this task, two agents collect subgoals together in a gridworld (Fig. \ref{gridworld}), and once all subgoals have been collected, either one of the agents needs to collect the final goal.

Both agents have four possible actions: move up, down, left or right. The actions of both agents are fed to the environment simultaneously at each time step. Each step causes a small negative reward (-0.01), and collecting goals results in a positive reward (0.05). An episode ends when the final goal is collected or if the maximum amount of steps have been taken. As our setting is fully cooperative, both agents share the same reward.

For this task, several populations of agents with various skill levels are trained using self-play and by collecting agents from different phases of the training.  Our solution tackles the non-stationarity problem \cite{hernandez2017survey,nguyen2020deep} by running the training in two phases. During the first phase (\textit{clone training}) the self-play training is conducted with the most recent versions of the policy. During the second phase, the partner is sampled from the older versions of the policy, but only the parameters of the latest version of the policy are updated. Thus, during the second phase, which we call \textit{co-op training}, only one agent keeps on learning, thus avoiding the non-stationarity problem. We produce populations with three different skill levels: fully trained (\textit{skilled}), untrained (\textit{novice}) and a skill level that reaches a specified intermediate performance (\textit{intermediate}).

The state $s_t$ from the environment consist of the gridworld coordinates of all objects and flags indicating whether certain subgoals are already collected. We use a modified relation network \cite{santoro2017simple} to produce improved representations  of the states in all networks during the training and execution. In this experiment, the different tasks refer to cooperating with partner agents with various skill levels.   

The skill prediction network uses a standard LSTM implementation. We minimize the negative log likelihood under the task distribution $\mathcal{L}^{XE} = \mathbb{E}_{p(M)}[  -log(\hat{T_i})]$, where $\hat{T_i}$ is the prediction of the correct task. The policy network is a feedforward network with PPO loss and with separate policy and value networks. More details of the implementation is available in the appendix.

\subsection{Travelling Salesman Gridworld (TSG) results}

In the TSG experiment, we compare the behaviour-conditioned policy against a baseline LSTM policy with a similar capacity which can learn the required functionality in an end-to-end manner, see the appendix for details. As with the matrix game experiment, the baseline LSTM policy and the behaviour conditioned policy receive the ground truth information about the partner agent behaviour and use it in the form of a centralized critic. All populations that are trained with five random seeds, are partnered during evaluation with agents from an additional reference population that is not seen during the training in order to penalize overfitting to the behaviours present within the training population.

The results in Table \ref{results_short} show that both the behaviour conditioned policy and baseline reach strong absolute performance with all partner types when comparing to the optimal policy. Furthermore, the performance of both methods consistently improves as the partner agent becomes more skilled. This shows that the meta-learning distribution generated through self-play not only produces strong co-operative agents, but also allows for gaining the ability to quickly adapt based on the behaviour of the partner agent in novel situations. The results also show that the behaviour conditioned policy outperforms the baseline for all partner types. This indicates that the inductive bias present for exploiting the two-fold structure of the task is useful for fast adaptation, and that the labels produced during self-play can be used as an additional training signal in parallel to their typical use for a centralized critic.

\begin{table}[h!]
\begin{center}
\begin{tabular}{ |l|c|c|c| } 
\hline

Method & partner type & Mean episode length & Mean return \\
 \hline
 
Behaviour-conditioned policy & skilled & 15.9 $\pm 0.1$ & 0.080 $\pm 0.001$ \\
Behaviour-conditioned policy  & intermediate  & 22.9 $\pm 0.1$ & 0.011 $\pm 0.001$ \\
Behaviour-conditioned policy & novice & 28.0 $\pm 0.4$ & -0.040 $\pm0.004$ \\
  \hline
 
LSTM policy & skilled & 17.8 $\pm 0.2$ & 0.062 $\pm 0.002$\\
LSTM policy& intermediate & 25.4 $\pm 0.8$ & -0.015 $\pm 0.009$\\
LSTM policy& novice &  31.8 $\pm 0.6$ & -0.084 $\pm 0.007$\\

  \hline
 Optimal & skilled & 12.9 & 0.121\\ 
 Optimal & novice & $26.2^{*}$ & $-0.012^{*}$ \\ 
 \hline
\end{tabular}
\end{center}
\caption{Main results for simulations in 11 x 11 gridworld, two agents, four subgoals and one final goal. The results show mean values and standard deviations over five random seeds. *This is optimal when one agent collects everything. In our simulations, the other novice agent can accidentally pick up subgoals or the final goal, meaning that the simulation can exceed this value.}
\label{results_short}
\end{table}

\section{Discussion}
In this paper, we introduced ways to improve cooperation with an unknown partner agent type by synthetically generating population of agents and training a meta-learner by using these populations. Furthermore, we suggested an architecture that exploits efficiently the two-fold structure that is common in many real world scenarios.

The results showed that self-play can be used to construct useful task distributions for meta-learning the ability to quickly adapt to different partner types in novel cooperative situations. The results additionally indicate that utilizing data produced by self-play together with suitable inductive biases further improves the performance. 

Our experiments covered cases where the latent variable that explained the behaviour of the partner agent was the skill level of the partner agent. Depending on the tasks, many other latent variables could be identified. An interesting research direction would be to explore methods for incorporating the identification of important factors as a part of the autocurriculum.

\section{Acknowledgements}

This work was supported by the Academy of Finland (Flagship programme: Finnish Center for Artificial Intelligence FCAI, grants 319264, 292334). 

\printbibliography

\section{Appendix: Details of the neural network implementations}

\subsection{Matrix game experiment}

The baseline RL2 implementation has an LSTM layer with 32 units, followed by a MLP layer with 16 units and a linear layer with softmax. The value network does not share any weights with the policy network, and it has an LSTM layer with 16 units, followed by a MLP layer of 16 units and a linear layer.

The behaviour-conditioned policy network has two MLP layers with 6 and 16 units followed by a linear layer with softmax. The value network does not share any weights with the policy network, and it has similar structure as the policy network, except there is no softmax.

The task prediction network has an LSTM layer with 32 units, followed by one MLP layer with 16 units and a linear layer with softmax.

\subsection{Travelling salesman gridworld experiment}

All networks use modified version of relation net \cite{santoro2017simple}, where the pairwise relations are processed only between the agents and other objects, instead of relations between all possible objects. The relation nets have 7 MLP layers with the hidden size of 128 before the summation operation, and 2 MLP layers of 64 hidden units after the summation. All LSTM layers have 64 hidden units. 

The LSTM policy network (baseline) implementation has an LSTM layer on top of the relation net and a softmax layer to produce action distribution. The value network is a separate network with 2 MLP layers on top of the relation net. The baseline was trained 3000 iterations of clone training, and another 1000 of co-op training, with a dataset batch size of 4000 time steps. 

The skill prediction network has an LSTM layer on top of the relation net and 2 MLP layers with softmax layer. It was trained with 5000 training iteration with a batch size of 4000 time steps. 

The behaviour-conditioned policy network has 2 MLP layers and a softmax on top of the relation net. It was trained 3000 iterations of clone training, and another 1000 of co-op training, with a dataset batch size of 4000 time steps. 

The optimizer is Adam with initial learning rate 5e-4 and scheduled learning rate decrease.

\end{document}